\documentclass[10pt,twocolumn,letterpaper]{article}

\usepackage{iccv}
\usepackage{times}
\usepackage{epsfig}
\usepackage{graphicx}
\usepackage{amsmath}
\usepackage{amssymb}
\usepackage{multirow}
\usepackage{tabularx}
\newcommand{\PreserveBackslash}[1]{\let\temp=\\#1\let\\=\temp}
\newcolumntype{C}[1]{>{\PreserveBackslash\centering}p{#1}}

\usepackage[pagebackref=true,breaklinks=true,letterpaper=true,colorlinks,bookmarks=false]{hyperref}

\iccvfinalcopy 

\def\iccvPaperID{251} 

\newcommand\blfootnote[1]{%
  \begingroup
  \renewcommand\thefootnote{}\footnote{#1}%
  \addtocounter{footnote}{-1}%
  \endgroup
}

\ificcvfinal\fi
\begin{document}

\title{Learning Lightweight Lane Detection CNNs by Self Attention Distillation}

\author{\textbf{Yuenan Hou}$^{1}$, \textbf{Zheng Ma}$^{2}$, \textbf{Chunxiao Liu}$^{2}$, \textbf{and Chen Change Loy}$^{3\dagger}$\\
$^{1}$The Chinese University of Hong Kong $^{2}$SenseTime Group Limited $^{3}$Nanyang Technological University\\
hy117@ie.cuhk.edu.hk, \{mazheng, liuchunxiao\}@sensetime.com, ccloy@ntu.edu.sg
}

\maketitle

\def\algorithmname{SAD}

\begin{abstract}

Training deep models for lane detection is challenging due to the very subtle and sparse supervisory signals inherent in lane annotations. Without learning from much richer context, these models often fail in challenging scenarios, e.g., severe occlusion, ambiguous lanes, and poor lighting conditions. In this paper, we present a novel knowledge distillation approach, i.e., Self Attention Distillation (\algorithmname), which allows a model to learn from itself and gains substantial improvement without any additional supervision or labels.
Specifically, we observe that attention maps extracted from a model trained to a reasonable level would encode rich contextual information. The valuable contextual information can be used as a form of `free' supervision for further representation learning through performing top-down and layer-wise attention distillation within the network itself. \algorithmname~can be easily incorporated in any feedforward convolutional neural networks (CNN) and does not increase the inference time.
We validate \algorithmname~on three popular lane detection benchmarks (TuSimple, CULane and BDD100K) using lightweight models such as ENet, ResNet-18 and ResNet-34. The lightest model, ENet-\algorithmname, performs comparatively or even surpasses existing algorithms. Notably, ENet-\algorithmname~has \textbf{20}~$\times$ fewer parameters and runs \textbf{10}~$\times$ faster compared to the state-of-the-art SCNN~\cite{pan2017spatial}, while still achieving compelling performance in all benchmarks. Our code is available at \url{https://github.com/cardwing/Codes-for-Lane-Detection}.


\end{abstract}

\section{Introduction}
\label{sec:introduction}

\begin{figure*}[t]
  \centering
  \includegraphics[width=0.78\linewidth]{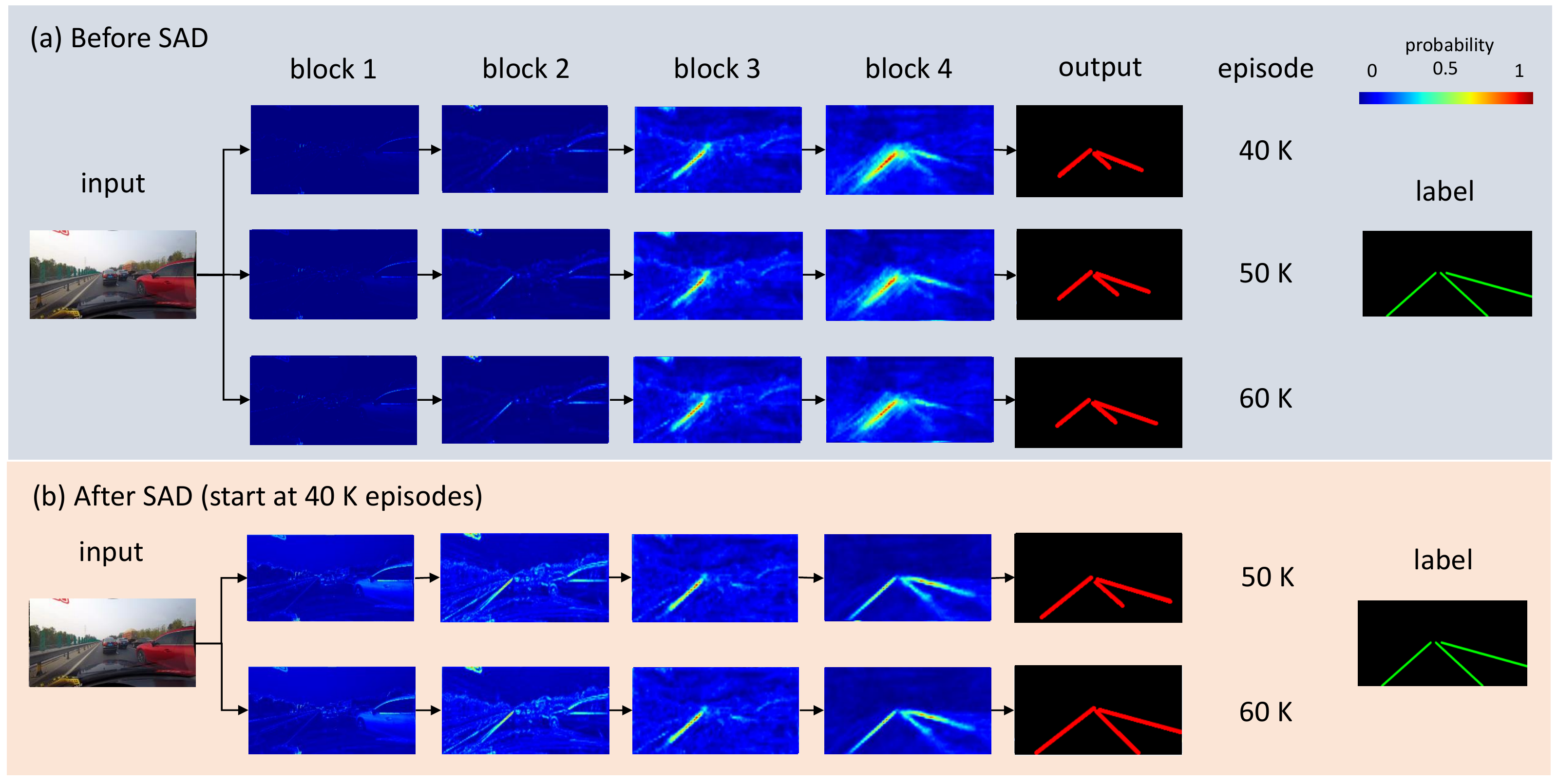}
  \vskip -0.2cm
  \caption{Attention maps of the ENet~\cite{paszke2016enet} before and after applying self attention distillation. Here, we extract the attention maps from the four stages/blocks following the design of ENet model. Note that self attention distillation is added in the 40 K episodes.}
  \centering
  \vskip -0.2cm
  \label{fig:attention_map}
\end{figure*}

\blfootnote{$\dagger$: Corresponding author.} 
Lane detection ~\cite{bertozzi1998gold} plays a pivotal role in autonomous driving as lanes could serve as significant cues for constraining the maneuver of vehicles on roads. Detecting lanes in-the-wild is challenging due to poor lighting conditions, occlusions caused by other vehicles, irrelevant road markings, and the inherent long and thin property of lanes. 

Contemporary algorithms ~\cite{chen2017rbnet, ghafoorian2018gan, lee2017vpgnet, pan2017spatial} typically adopt a dense prediction formulation, \ie, treat lane detection as a semantic segmentation task, where each pixel in an image is assigned with a binary label to indicate whether it belongs to a lane or not. These methods heavily rely on the segmentation maps of lanes as the supervisory signals. Since lanes are long and thin, the number of annotated lane pixels is far fewer than the background pixels. Learning from such subtle and sparse annotations becomes a major challenge in training deep models for the task. A plausible way is to increase the width of lane annotations. However, it may degrade the detection performance.   

Several schemes have been proposed to relieve the reliance of deep models on the sparse annotations, \eg, multi-task learning (MTL) and message passing (MP). For example, Lee \etal ~\cite{lee2017vpgnet} exploit vanishing points to guide the training of deep models and Pan \etal ~\cite{pan2017spatial} incorporate spatial MP in their lane detection models. MTL can indeed provide additional supervisory signals but it requires additional efforts, usually with human intervention, to prepare the annotations, \eg, scene segmentation maps, vanishing points, or drivable areas. MP can help propagate the information between neurons to counter the effect of sparse supervision and better capture the scene context. However, it increases the inference time significantly due to the overhead of MP. For instance, applying MP in a layer of SCNN~\cite{pan2017spatial} contributes \textbf{35}\% of its total feed-forward time.   

In this work, we present a simple yet novel approach that allows a lane detection network to reinforce representation learning of itself without the need of additional labels and external supervisions. In addition, it does not increase the inference time of the base model. 
Our approach is named \textit{Self-Attention Distillation} (\algorithmname). 
As the name implies, \algorithmname~allows a network to exploit attention maps derived from its own layers as the distillation targets for its lower layers. Such an attention distillation mechanism is used to complement the usual segmentation-based supervised learning.

\algorithmname~is motivated by an interesting observation --  when a lane detection network is trained to a reasonable level, attention maps derived from different layers would capture diverse and rich contextual information that hints the lane locations and a rough outline of the scene, as shown in Fig.~\ref{fig:attention_map} (before SAD at 40K episodes).
By adding \algorithmname~to the learning of this half-trained model, \ie, having the preceding block to mimic the attention maps of a deeper block, \eg, block~3 $ \xrightarrow[]{\text{mimic}}$ block~4 and block~ 2 $ \xrightarrow[]{\text{mimic}}$ block~3, the network can learn to strengthen its representations, as shown in Fig.~\ref{fig:attention_map} (after SAD):
(1) the attention maps of lower layers are refined, with richer scene contexts captured by the visual attention, and (2) the better representation learned at lower layers in turn benefits the deeper layers. For instance, although block 4 does not learn from any distillation targets, its representation is reinforced, as evident from the much distinct attention at the lane locations. 
By contrast, without using SAD, the visual attentions of different layers of the same network hardly improve despite continual training up to 60K episodes.

\algorithmname~opens a new possibility of training accurate lane detection networks apart from deploying existing techniques such as multi-task learning and message passing, which can be expensive. It allows us to train small networks with excellent visual attention that is on par with very deep networks. In our experiments, we successfully demonstrate the effectiveness of \algorithmname~on a few popular lightweight models, \eg, ENet~\cite{paszke2016enet}, ResNet-18~\cite{he2016deep} and ResNet-34~\cite{he2016deep}.

In summary, our contributions are three-fold:
(1) We propose a novel attention distillation approach, i.e., \algorithmname, to enhance the representation learning of CNN-based lane detection models. \algorithmname~is only used in the training phase and brings no computational cost during the deployment. Our work is the first attempt of using a network's own attention maps as the distillation targets.
(2) We carefully and systematically investigate the inner mechanism of \algorithmname, the consideration of choosing among different layer-wise mimicking paths, and the timepoint of introducing \algorithmname~to the training process for improved gains.
(3) We verify the usefulness of \algorithmname~on boosting the performance of small lane detection networks. We further present several architectural reformulations to ENet~\cite{paszke2016enet} for improved performance. 
Our lightweight model, ENet-\algorithmname, achieves state-of-the-art lane detection performance on TuSimple~\cite{tusimple}, CULane~\cite{pan2017spatial} and BDD100K~\cite{yu2018bdd100k}. It can serve as a strong backbone to facilitate future research on lane detection. 

\section{Related Work}
\label{sec:relatedwork}

\noindent \textbf{Lane detection.} Lane detection is conventionally handled via using specialized and hand-crafted features to obtain lane segments. These segments are further grouped to get the final results ~\cite{borkar2012novel, deusch2012random}. These methods have many shortcomings, e.g., requiring complex feature selection process, being lack of robustness and only applicable to relatively easy driving scenarios. 

Recently, deep learning has been employed to omit hand-crafted features altogether and learn to extract features in an end-to-end manner ~\cite{lee2017vpgnet, pan2017spatial, ghafoorian2018gan, chen2017rbnet}. These approaches usually adopt the dense prediction formulation, i.e., treat lane detection as a semantic segmentation task, where each pixel in an image is assigned with a label to indicate whether it belongs to a lane or not. For example, He \etal ~\cite{he2016accurate} propose Dual-View CNN (DVCNN) to handle lane detection. The method takes front-view and top-view images as inputs. Another popular paradigm performs lane detection from the perspective of instance segmentation. For instance, Neven \etal ~\cite{neven2018towards} divide lane detection into two stages. Specifically, they first perform binary segmentation that differentiates lane pixels and background pixels. These lane pixels are then classified into different lane instances. 

Several schemes have been proposed to complement the lane-based supervision and to capture richer scene context, \eg, multi-task learning and message passing. For example, Zhang \etal~\cite{zhang2018geometric} establish a framework that accomplishes lane boundary segmentation and road area segmentation simultaneously. Geometric constraints that lane boundaries and lane areas constitute the road are also included to further enhance the final performance. Mohsen \etal~\cite{ghafoorian2018gan} take lane labels as extra inputs and integrate generative adversarial network (GAN) into the original framework so that the segmentation maps resemble labels more. Pan \etal ~\cite{pan2017spatial} perform sequential massage passing between the outputs of top-level layers to better exploit the structural information. While the aforementioned methods do bring additional gains to the performance, multi-task learning requires extra annotations and message passing is not efficient since it propagates information in a sequential way. On the contrary, the proposed \algorithmname~is free from the requirement of extra annotations and it does not increase the inference time.  

\noindent \textbf{Knowledge and attention distillation.} Knowledge distillation was originally proposed by \cite{hinton2015distilling} to transfer the knowledge from large networks to small networks. Commonly in knowledge distillation, a small student network mimics the intermediate outputs of large teacher networks as well as the labels. In \cite{furlanello2018born,yim2017gift} the student and teacher networks share the same capacity and mimicking is performed between pairs of layers with same dimensionality. Hou \etal ~\cite{hou2018learning} also investigate knowledge distillation performed between heterogeneous networks. Recent studies \cite{zagoruyko2016paying, Wang_2017_CVPR} have expanded knowledge distillation to attention distillation. For instance, Sergey \etal ~\cite{zagoruyko2016paying} introduce two types of attention distillation, \ie, activation-based attention distillation and gradient-based attention distillation. In both kinds of distillation, a student network is trained through learning attention maps derived from a teacher network. 
%
%
The proposed \algorithmname~differs to \cite{zagoruyko2016paying} in that our method does not need a teacher network. Distillation is conducted in a layer-wise and top-down manner, in which attention knowledge is propagated layer by layer. This is new in the literature. 
It is noteworthy that our focus is to investigate the possibility of distilling layer-wise attention for self-learning. This differs from existing studies on using visual attention for weighting features~\cite{chen2016attention, jaderberg2015spatial, zagoruyko2016paying}.

\section{Methodology}
\label{sec:methodology}


Lane detection is commonly formulated as a semantic segmentation task. More specifically, given an input image \textbf{X}, the objective is to assign a label $l_{ij}$ ($l_{ij}$ = 1, ..., $N_{c}$) to each pixel $(i, j)$ of \textbf{X}, comprising the segmentation map $s$. Here, $N_{c}$ is the number of classes. The objective is to learn a mapping $\mathcal{F}$: \textbf{X} $\mapsto$ $s$. Recent studies use CNN as $\mathcal{F}$ for end-to-end prediction. The task of lane existence prediction is also introduced to facilitate the evaluation process. We use $b$ to represent the binary labels that indicate the existence of lanes. Then, the function becomes $\mathcal{F}$: \textbf{X} $\mapsto$ ($s$, $b$). 


\subsection{Self Attention Distillation}

Apart from training our lane detection network with the aforementioned semantic segmentation and lane existence prediction losses, we aim to perform layer-wise and top-down attention distillation to enhance the representation learning process. The proposed SAD does not require any external supervision or additional labels since attention maps are derived from the network itself.

In general, attention maps can be divided into two categories, \ie, activation-based attention maps~\cite{zagoruyko2016paying} and gradient-based attention maps~\cite{zagoruyko2016paying}. The activation-based attention maps are obtained via processing the activation output of a specific layer while the gradient-based ones are obtained via using the layer's gradient output. In the experiment, we empirically find that activation-based attention distillation yields considerable performance gains while gradient-based attention distillation barely works. Hence, in the following sections we only discuss the activation-based attention distillation.

\noindent \textbf{Activation-based attention distillation.} We use $A_{m} \in R^{C_{m} \times H_{m} \times W_{m}}$ to denote the activation output of the $m$-th layer of the network, where $C_{m}$, $H_{m}$ and $W_{m}$ denote the channel, height and width, respectively. Let $M$ denote the number of layers in the network. The generation of the attention map is equivalent to finding a mapping function $\mathcal{G}$: $R^{C_{m} \times H_{m} \times W_{m}} \to R^{H_{m} \times W_{m}}$. The absolute value of each element in this map represents the importance of this element on the final output. Therefore, this mapping function can be constructed via computing statistics of these values across the channel dimension. More specifically, the following three operations~\cite{zagoruyko2016paying} can serve as the mapping function: $\mathcal{G}_{sum}(A_{m}) = \sum_{i=1}^{C_{m}}| A_{mi}|$, $\mathcal{G}_{sum}^{p}(A_{m}) = \sum_{i=1}^{C_{m}}| A_{mi}|^{p}$ and $\mathcal{G}_{\max}^{p}(A_{m}) = \max_{i=1, C_{m}}| A_{mi}|^{p}$. Here, $p > 1$ and $A_{mi}$ denotes the $i$-th slice of $A_{m}$ in the channel dimension.

\begin{figure}[b]
  \centering
  \includegraphics[width=1.0\linewidth]{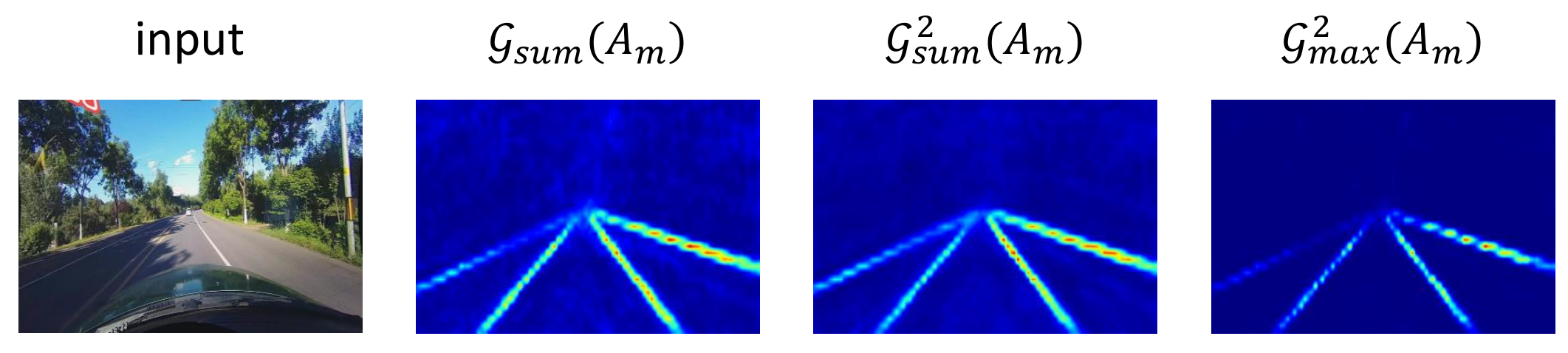}
  \caption{Attention maps of the block 4 of the ENet model using different mapping functions.}
  \centering
  \vskip -0.2cm
  \label{fig:func}
\end{figure}

\begin{figure*}[t]
  \centering
  \includegraphics[width=0.78\linewidth]{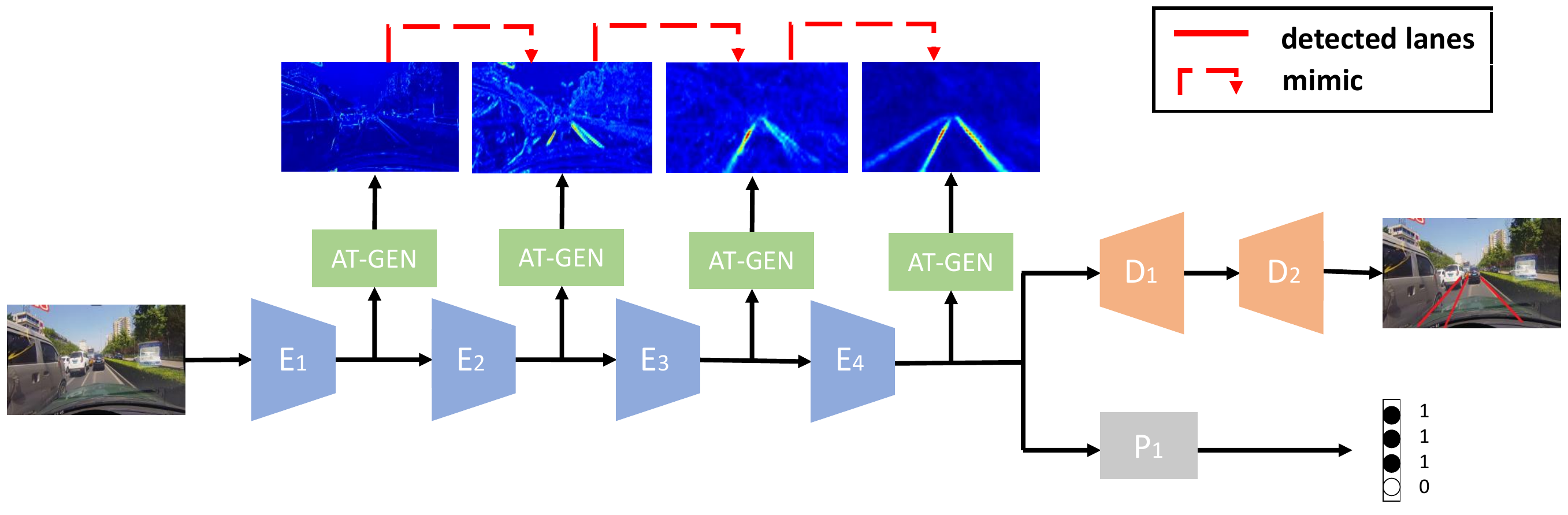}
  \vskip -0.2cm
  \caption{An instantiation of using \algorithmname. $E_{1}$ $\sim$ $E_{4}$ comprise the encoder of ENet~\cite{paszke2016enet}, $D_{1}$ and $D_{2}$ comprise the decoder of ENet. Following \cite{pan2017spatial}, we add a small network to predict the existence of lanes, denoted as $P_{1}$. AT-GEN is the attention generator.}
  \centering
  \vskip -0.2cm
  \label{fig:pipeline}
\end{figure*}

The differences between these mapping functions are depicted in Fig.~\ref{fig:func}. Compared with $\mathcal{G}_{sum}(A_{m})$, $\mathcal{G}_{sum}^{p}(A_{m})$ puts more weights to areas with higher activations. The larger the $p$ is, the more focus is placed on these highly activated areas. Compared with $\mathcal{G}_{\max}^{p}(A_{m})$, $\mathcal{G}_{sum}^{p}(A_{m})$ is less biased since it calculates weights across multiple neurons instead of selecting the maximum value of these neuron activations as the weight. In the experiment, we empirically find that using $\mathcal{G}_{sum}^{2}(.)$ as the mapping function yields the most performance gains.

\noindent \textbf{Adding SAD to training.} The intuition behind \algorithmname~is that the attention maps of previous layers can distil useful contextual information from those of successive layers. Following ~\cite{zagoruyko2016paying}, we also perform spatial softmax operation $\Phi(.)$ on $\mathcal{G}_{sum}^{2}(A_{m})$. Bilinear upsampling $\mathcal{B}(.)$ is added before the softmax operation if the size of original attention maps is different from that of targets. However, different from Sergey \etal ~\cite{zagoruyko2016paying} who perform attention distillation within two networks, the proposed self attention distillation is performed within the network itself. 

Adding SAD to an existing network is straight-forward. It is possible to introduce SAD at different timepoint of the training, which could affect the convergence time. We will show an evaluation in the experiment section. Here we assume an ENet half-trained to 40K episodes.
As shown in Fig. \ref{fig:pipeline}, we add an attention generator, abbreviated as AT-GEN, after each $E_2$, $E_3$, and $E_4$ encoder block of ENet. Formally, AT-GEN is represented by a function $\Psi(.) = \Phi(\mathcal{B}(\mathcal{G}_{sum}^{2}(.)))$. A successive layer-wise distillation loss is formulated as follows:
\begin{equation}
\label{eqn:sad_loss}
\begin{split}
\mathcal{L}_\mathrm{distill}(A_{m}, A_{m+1}) =
\sum_{m=1}^{M-1} \mathcal{L}_\mathrm{d} (\Psi(A_{m}), \Psi(A_{m+1})),
\end{split}
\end{equation}  
where $\mathcal{L}_\mathrm{d}$ is typically defined as a $L_{2}$ loss and $\Psi(A_{m+1})$ is the target of the distillation loss. In the example shown in Fig. \ref{fig:pipeline}, we have the number of layers $M = 4$. Note that we do not assign different weights to different SAD paths, although this is possible. We found that this uniform scheme works well in our experiments. 

The total loss is comprised of four terms:
%
\begin{equation}
\label{eqn:total_loss}
\begin{split}
\mathcal{L} = & \underbrace{\mathcal{L}_\mathrm{seg}(s, \hat{s}) + \alpha \mathcal{L}_\mathrm{IoU}(s, \hat{s})}_{\text{segmentation loss}} \\
& + \underbrace{\beta \mathcal{L}_\mathrm{exist}(b, \hat{b})}_{\text{existence loss}} + \underbrace{\gamma \mathcal{L}_\mathrm{distill}(A_{m}, A_{m+1})}_{\text{distillation loss}}.
\end{split}
\end{equation}
Here, the first two terms are segmentation losses that comprise of the standard cross entropy loss $\mathcal{L}_\mathrm{seg}$ and the IoU loss $\mathcal{L}_\mathrm{IoU}$. The IoU loss aims at increasing the intersection-over-union between the predicted lane pixels and ground-truth lane pixels. It is formulated as $\mathcal{L}_{IoU} = 1 - \frac{N_{p}}{N_{p} + N_{g} - N_{o}}$, where $N_{p}$ is the number of predicted lane pixels, $N_{g}$ is the number of ground-truth lane pixels and $N_{o}$ is the number of lane pixels in the overlapped areas between predicted lane areas and ground-truth lane areas.
$\mathcal{L}_\mathrm{exist}$ is the binary cross entropy loss. $\hat{s}$ is the segmentation map produced by the network and $\hat{b}$ is the prediction of the existence of lanes. The parameters $\alpha$, $\beta$, and $\gamma$ balance the influence of segmentation losses, existence loss, and distillation loss on the final task. 


It is noteworthy that the SAD paths can be generalized to dense connections beyond the example shown here. For instance, we can add block~1 $ \xrightarrow[]{\text{mimic}}$ block~3, block~1 $ \xrightarrow[]{\text{mimic}}$ block~4, and block~ 2 $ \xrightarrow[]{\text{mimic}}$ block~4 in addition to the current paths. In general, the number of possible SAD paths for a network with a depth of $M$ layers is $\frac{M(M-1)}{2}$. We will evaluate this possibility in our experiments. 

\begin{figure*}
  \centering
  \includegraphics[width=0.75\linewidth]{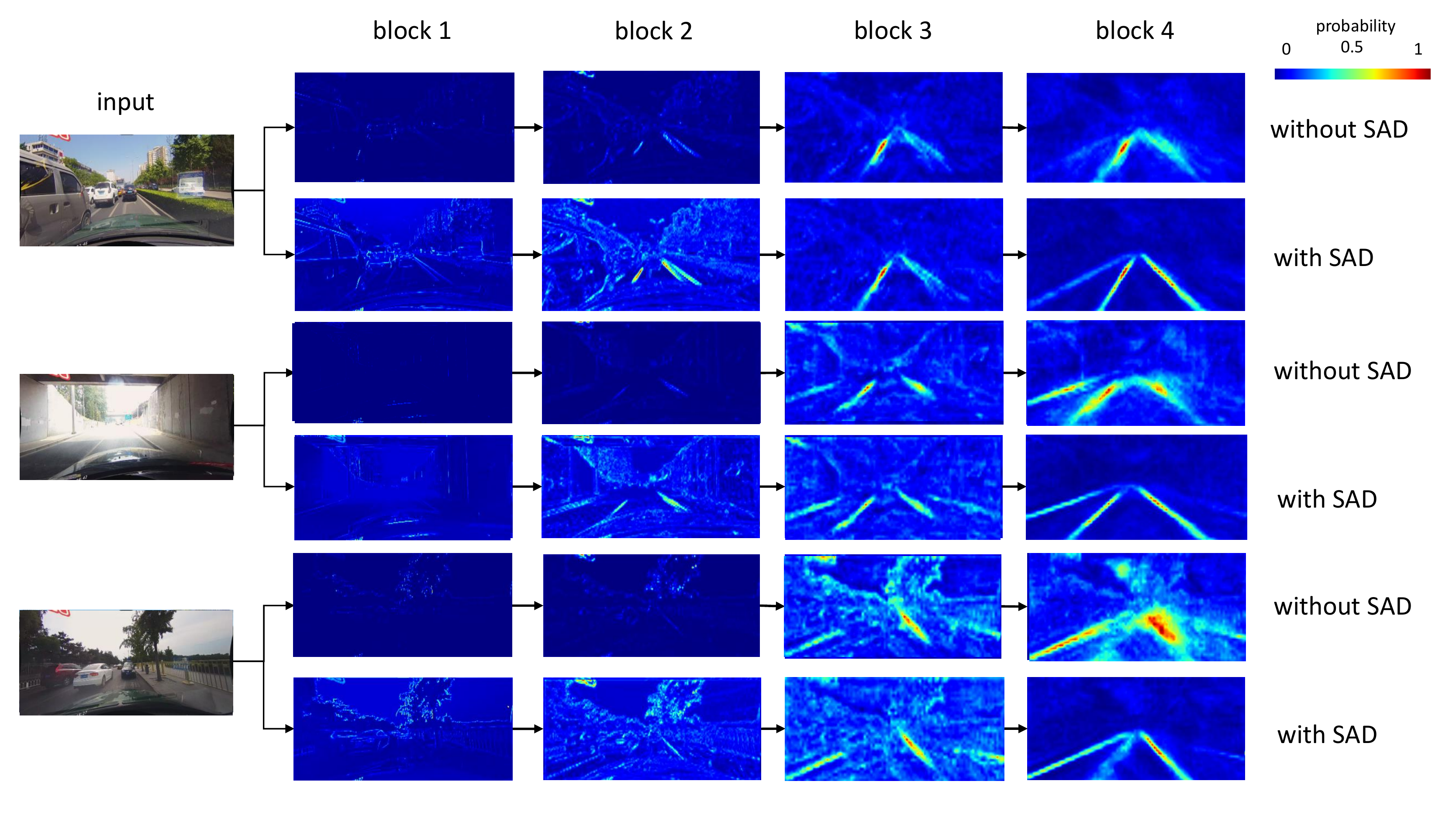}
  \caption{Attention maps of ENet with and without self attention distillation. Both networks with and without SAD are trained up to 60K episodes. SAD is applied to ENet at 40K training episodes.}
  \centering
  \vskip -0.2cm
  \label{fig:learning}
\end{figure*}

\noindent \textbf{Visualization of attention maps with and without SAD.} 
We investigate the influence of SAD by studying the attention maps of different blocks in ENet with and without SAD. More results will be reported in Section~\ref{sec:experiments}. Both networks with and without SAD are trained up to 60K episodes. 
We visualize the attention maps of four existing blocks in ENet. As can be observed in Fig.~\ref{fig:learning}, after adding \algorithmname, the attention maps of ENet become more concentrated on task-relevant objects, \eg, lanes, vehicles and road curbs. This would in turn improve the lane detection accuracy, as we will show in the experiments.


\subsection{Lane Prediction}
\label{subsec:inference}

The output of the model is not post-processed for TuSimple and BDD100K except CULane. For CULane, in the inference stage, we feed the image into the ENet model. Then the multi-channel probability maps and the lane existence vector are obtained. Following~\cite{pan2017spatial}, the final output is obtained as follows: First, we use a 9 $\times$ 9 kernel to smooth the probability maps. Then, for each lane whose existence probability is larger than 0.5, we search the corresponding probability map every 20 rows for the position with the highest probability value. In the end, we use cubic splines to connect these positions to get the final output.

\subsection{Architecture Design}
\label{subsec:architecture}

The original ENet model is an encoder-decoder structure comprised of $E_{1} \sim E_{4}$, $D_{1}$ and $D_{2}$. Following \cite{pan2017spatial}, we add a small network $P_{1}$ to predict the existence of lanes. The encoder module is shared to save memory space.
Apart from this modification, we also observed some useful techniques to modify ENet for achieving better performance in the lane detection task.
Dilated convolution \cite{yu2015multi} is added to replace the original convolution layers in the lane existence prediction branch to increase the receptive field of the network without increasing the number of parameters. In the original design, the resolution of feature maps of $E_{4}$ is only 36 $\times$ 100 for CULane. This leads to severe loss of information. Hence, we use feature concatenation to fuse the output of $E_{4}$ with that of $E_{3}$ so that the output of the encoder can benefit from information encoded in previous layers. 

\section{Experiments}
\label{sec:experiments}



\begin{table*}[!t]
\caption{Basic information of three lane detection datasets.}
\vskip 0.1cm
\label{dataset_table}
\centering
\small{
\begin{tabular}{c|c|c|c|c|c|c|c}
\hline
Name & \# Frame & Train & Validation & Test & Resolution & Road Type & \# Lane $>$ 5 ? \\
\hline
\hline
TuSimple~\cite{tusimple} & 6, 408 & 3, 268 & 358 & 2, 782 & 1280 $\times$ 720 & highway & $\times$ \\
CULane~\cite{pan2017spatial} & 133, 235 & 88, 880 & 9, 675 & 34, 680 & 1640 $\times$ 590 & urban, rural and highway & $\times$ \\
BDD100K~\cite{yu2018bdd100k} & 80, 000 & 60, 000 & 10, 000 & 10, 000 & 1280 $\times$ 720 & urban, rural and highway &  $\surd$ \\
\hline
\end{tabular}
}
\vspace{-3ex}
\end{table*}

\begin{figure}[t]
  \centering
  \includegraphics[width=1.0\linewidth]{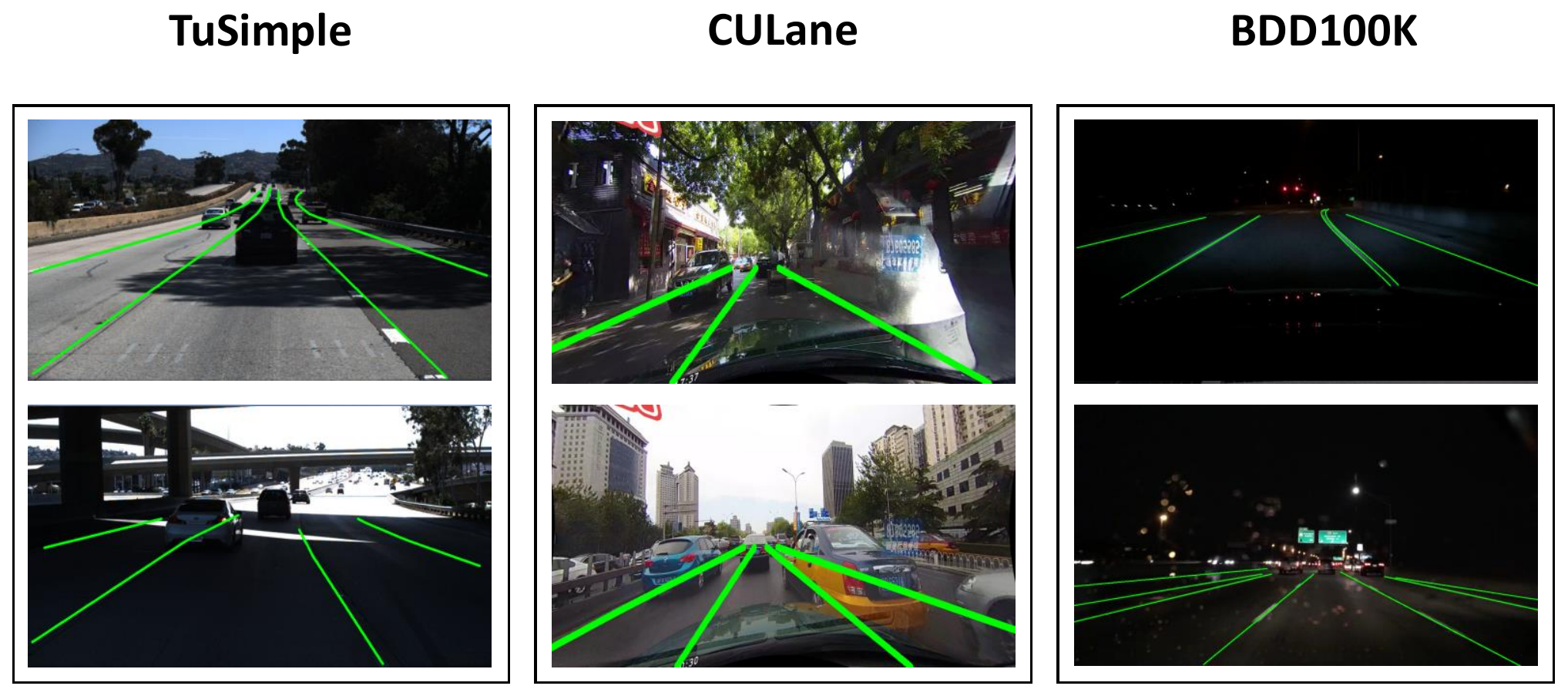}
  \vskip -0.2cm
  \caption{Typical video frames of TuSimple, CULane and BDD100K datasets. Ground-truth lanes are marked in green color.}
  \centering
  \vskip -0.2cm
  \label{fig:dataset}
\end{figure}

\vspace{0.1cm}
\noindent
\textbf{Datasets}.
Figure~\ref{fig:dataset} shows several video frames of three datasets that we use in our experiments. They are TuSimple~\cite{tusimple}, CULane~\cite{pan2017spatial} and BDD100K~\cite{yu2018bdd100k}. TuSimple and CULane are widely used in the literature. Many algorithms~\cite{pan2017spatial, neven2018towards, ghafoorian2018gan} have been tested in TuSimple since it was the largest lane detection dataset before 2018. As to CULane, it contains many challenging driving scenarios like crowded road conditions or roads under poor lighting (see Fig.~\ref{fig:dataset}). BDD100K is originally designed for lane instance classification. However, since there are typically multiple lanes in an image and these lanes are usually very close to each other, using instance segmentation algorithms will yield inferior performance. Therefore, we choose to only detect lanes without differentiating lane instances for BDD100K. 
We discuss the details of transforming the original ground truths for our task in the following section on implementation details. 
Table \ref{dataset_table} summarizes their details. Note that the last column of Table~\ref{dataset_table} shows that TuSimple and CULane have no more than 5 lanes in a video frame while BDD100K typically contains more than 8 lanes in a video frame. Besides, TuSimple is relatively easy while CULane and BDD100K are more challenging considering the total number of video frames and road types. Note that the original BDD100K dataset provides 100K video frames, in which 70K are used for training, 10K for validation and 20K for testing. However, since the ground-truth labels of the testing partition are not publicly available, we keep the training set unchanged but use the original validation set for testing. 
A new validation set is allocated separately from the training set, as shown in Table~\ref{dataset_table}.

\noindent
\textbf{Evaluation metrics}. To facilitate comparisons against previous studies, we follow the literature and use the corresponding evaluation metrics for each particular dataset.  

\noindent \textit{1) TuSimple.} We use the official metric (accuracy) as the evaluation criterion. Besides, false positive ($FP$) and false negative ($FN$) are also reported. Accuracy is computed as~\cite{tusimple}: $Accuracy = \frac{N_{pred}}{N_{gt}}$,
%
%
where $N_{pred}$ is the number of correctly predicted lane points and $N_{gt}$ is the number of ground-truth lane points.

\noindent \textit{2) CULane.} Following~\cite{pan2017spatial}, to judge whether a lane is correctly detected, we treat each lane as a line with 30 pixel width and compute the intersection-over-union (IoU) between labels and predictions. Predictions whose IoUs are larger than 0.5 are considered as true positives (TP). Then, we use $F_{1}$ measure as the evaluation metric, which is defined as: $F_{1} = \frac{2 \times Precision \times Recall}{Precision + Recall}$,
%
%
where $Precision = \frac{TP}{TP + FP}$ and $Recall = \frac{TP}{TP + FN}$.

\noindent \textit{3) BDD100K.} Since there are typically more than 8 lanes in an image, we decide to use pixel accuracy and IoU of lanes to evaluate the performance of different models. 

\noindent
\textbf{Implementation details}.
Following~\cite{pan2017spatial}, we resize the images of TuSimple and CULane to 368$\times$640 and 288$\times$800, respectively. As to BDD100K, we resize the image to 360$\times$640 to save memory usage. The lanes of BDD100K are labelled by two lines. Training the networks using the provided labels is tricky. Therefore, based on these two lines, we calculate the center lines as new targets. We dilate ground-truth lanes of the training set of BDD100K as 8 pixels to provide denser targets while keeping these of testing set unchanged (2 pixels). We use SGD~\cite{bottou2010large} to train our models and the learning rate is set to 0.01. Batch size is set as 12 and the total number of training episodes is set as 1800 for TuSimple and 60K for CULane and BDD100K. The cross entropy loss of background pixels is multiplied by 0.4. Loss coefficients $\alpha$, $\beta$, and $\gamma$ are set as 0.1. Since we select lane pixel accuracy and IoU as the evaluation criterion for BDD100K dataset, we alter the original segmentation branch to output binary segmentation maps to facilitate the evaluation on BDD100K. The lane existence prediction branch is also removed for the BDD100K evaluation.

We empirically found that several practical techniques, \ie, data augmentation and IoU loss, can considerably enhance the performance of CNN-based lane detection models. As to data augmentation, we use random rotation, random cropping and horizontal flipping to process the input images.  
In our experiments, we apply the same segmentation losses and augmentation strategy to our method, SCNN, ResNet baselines, and deep supervision methods, to ensure a fair comparison. Since the source codes of LaneNet~\cite{neven2018towards} and EL-GAN~\cite{ghafoorian2018gan} are not available, we use their results reported in their papers.

\subsection{Results}

\begin{table}[!t]
\caption{Performance of different algorithms on TuSimple testing set. Here "R-18-\algorithmname~" denotes ResNet-18 + \algorithmname~and we use the same abbreviation in the following sections.}
\label{tusimple_table}
\centering
\small{
\begin{tabular}{c|c|c|c}
\hline
Algorithm & Accuracy & FP & FN \\
\hline \hline
ResNet-18~\cite{he2016deep} & 92.69\% & 0.0948 & 0.0822 \\
ResNet-34~\cite{he2016deep} & 92.84\% & 0.0918 & 0.0796 \\
ENet~\cite{paszke2016enet} & 93.02\% & 0.0886 & 0.0734 \\
LaneNet~\cite{neven2018towards} & 96.38\% & 0.0780 & 0.0244 \\
EL-GAN~\cite{ghafoorian2018gan} & 96.39\% & \textbf{0.0412} & 0.0336 \\
SCNN~\cite{pan2017spatial} & 96.53\% & 0.0617 & \textbf{0.0180} \\
\hline \hline
\textbf{R-18-\algorithmname~(ours)} & 96.02\% & 0.0786 & 0.0451 \\
\textbf{R-34-\algorithmname~(ours)} & 96.24\% & 0.0712 & 0.0344 \\
\textbf{ENet-\algorithmname~(ours)} & \textbf{96.64\%} & 0.0602 & 0.0205 \\
\hline
\end{tabular}
}
\vspace{-2ex}
\end{table}

\begin{table*}[!t]
\caption{Performance ($F_{1}$-measure) of different algorithms on CULane testing set. For crossroad, only FP is shown. The second column denotes the proportion of each scenario in the testing set.}
\label{culane_table}
\centering
\small{
\begin{tabular}{c|c|c|c|c|c||c|c}
\hline
Category & Proportion & \textbf{ENet-\algorithmname~} & \textbf{R-18-\algorithmname~} & \textbf{R-34-\algorithmname~} & \textbf{R-101-\algorithmname~} & ResNet-101~\cite{he2016deep} & SCNN~\cite{pan2017spatial} \\
\hline \hline
Normal & 27.7\% & 90.1 & 89.8 & 89.9 & \textbf{90.7} & 90.2 & 90.6 \\
Crowded & 23.4\% & 68.8 & 68.1 & 68.5 & \textbf{70.0} & 68.2 & 69.7 \\
Night & 20.3\% & 66.0 & 64.2 & 64.6 & \textbf{66.3} & 65.9 & 66.1 \\
No line & 11.7\% & 41.6 & 42.5 & 42.2 & \textbf{43.5} & 41.7 & 43.4 \\
Shadow & 2.7\% & 65.9 & 67.5 & \textbf{67.7} & 67.0 & 64.6 & 66.9 \\
Arrow & 2.6\% & 84.0 & 83.9 & 83.8 & \textbf{84.4} & 84.0 & 84.1 \\
Dazzle light & 1.4\% & \textbf{60.2} & 59.8 & 59.9 & 59.9 & 59.8 & 58.5 \\
Curve & 1.2\% & 65.7 & 65.5 & \textbf{66.0} & 65.7 & 65.5 & 64.4 \\
Crossroad & 9.0\% & 1998 & 1995 & \textbf{1960} & 2052 & 2183 & 1990 \\
Total & -- & 70.8 & 70.5 & 70.7 & \textbf{71.8} & 70.8 & 71.6 \\
\hline \hline
Runtime (ms) & -- & \textbf{13.4} & 25.3 & 50.5 & 171.2 & 171.2 & 133.5  \\
Parameter (M) & -- & \textbf{0.98} & 12.41 & 22.72 & 52.53 & 52.53 & 20.72  \\
\hline
\end{tabular}
}
\vspace{-3ex}
\end{table*}

\begin{table}[!t]
\caption{Comparative results on BDD100K test set.}
\label{bdd100k_table}
\centering
\small{
\begin{tabular}{c|c|c}
\hline
Algorithm & Accuracy & IoU \\
\hline \hline
ResNet-18~\cite{he2016deep} & 30.66\% & 11.07 \\
ResNet-34~\cite{he2016deep} & 30.92\% & 12.24 \\
ResNet-101~\cite{he2016deep} & 34.45\% & 15.02 \\
ENet~\cite{paszke2016enet} & 34.12\% & 14.64 \\
SCNN~\cite{pan2017spatial} & 35.79\% & 15.84 \\
\hline \hline
\textbf{R-18-\algorithmname~(ours)} & 31.10\% & 13.29 \\
\textbf{R-34-\algorithmname~(ours)} & 32.68\% & 14.56 \\
\textbf{R-101-\algorithmname~(ours)} & 35.56\% & 15.96 \\
\textbf{ENet-\algorithmname~(ours)} & \textbf{36.56\%} & \textbf{16.02} \\
\hline
\end{tabular}
}
\vspace{-2ex}
\end{table}

Tables \ref{tusimple_table}-\ref{bdd100k_table} summarize the performance of our methods, \ie, ResNet-18-\algorithmname, ResNet-34-\algorithmname, and ENet-\algorithmname~against state-of-the-art algorithms on the testing set of TuSimple, CULane and BDD100K datasets.
We also report the runtime and parameter count of different algorithm in Table~\ref{culane_table} so that we can compare the performance with the complexity of the model taken into account. The runtime is recorded using a single GPU (GeForce GTX TITAN X) and the final value of runtime is obtained after averaging the runtime of 100 samples. 

It is observed that ENet-\algorithmname~outperforms all baselines in BDD100K while achieving compelling performance in TuSimple and CULane. Considering that ENet-\algorithmname~has 20 $\times$ fewer parameters and runs 10 $\times$ faster compared with SCNN on CULane testing set, the performance strongly suggests the effectiveness of \algorithmname. 
It is observed that ResNet-18-SAD and ResNet-34-SAD achieve slightly inferior performance to ENet-\algorithmname~despite their larger model capacity. The is because ResNet-18 and ResNet-34 only use spatial upsampling as the decoder while ENet has a specially designed decoder for the task. 
It is noteworthy that SAD also helps given a deeper model. Specifically, we apply \algorithmname~to ResNet-101, and find that it increases the F$_{1}$-measure from 70.8 to 71.8 in CULane and the accuracy increases from 34.45\% to 35.56\% in BDD100K. 

\begin{figure}[t]
  \centering
  \includegraphics[width=\linewidth]{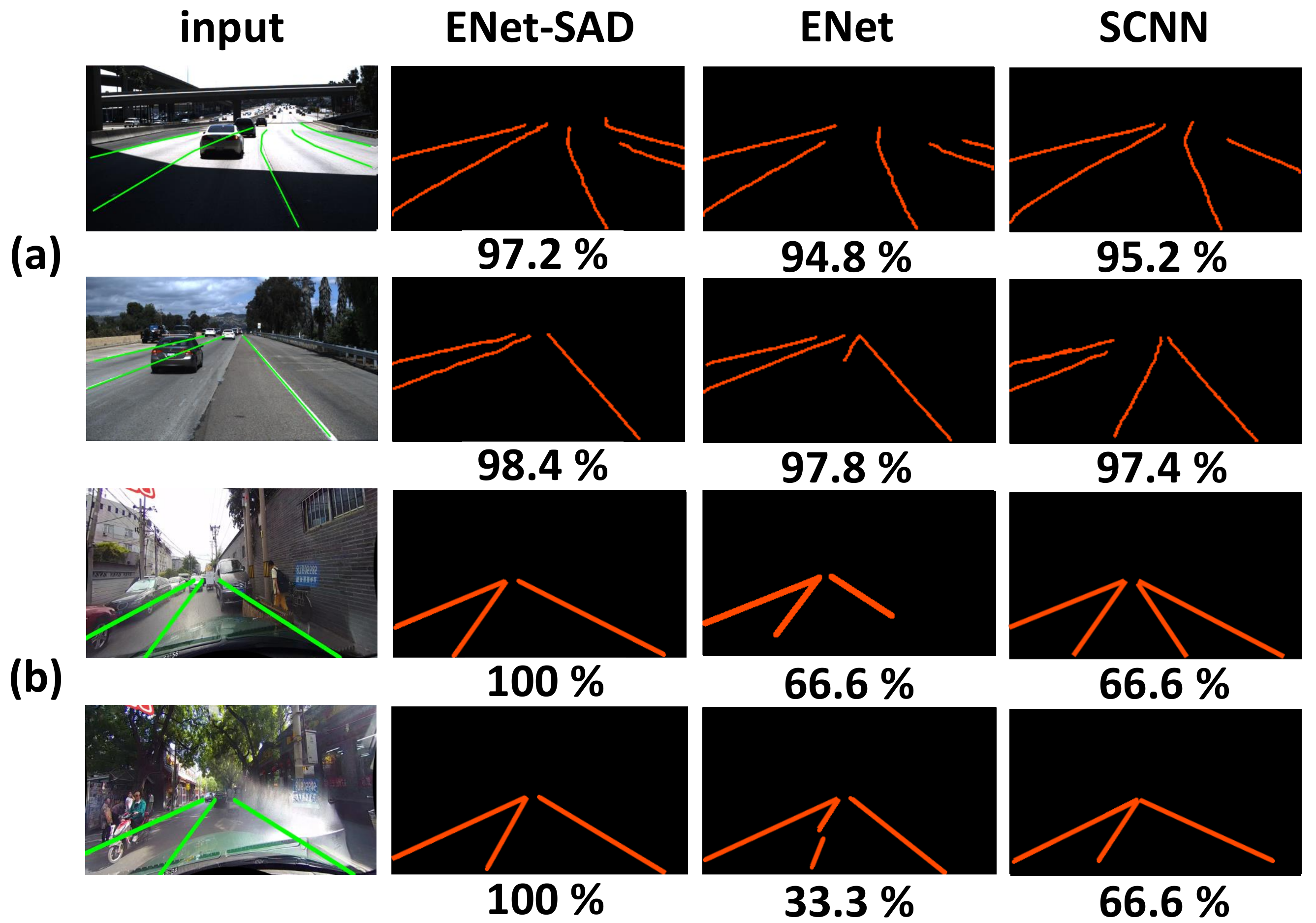}
  \vskip -0.1cm
  \caption{Performance of different algorithms on (a) TuSimple and (b) CULane testing sets. The number below each image denotes the accuracy. Ground-truth lanes are drawn on the input image.}
  \centering
  \vskip -0.25cm
  \label{fig:tusimple_result}
\end{figure}

\begin{figure}[t]
  \centering
  \includegraphics[width=\linewidth]{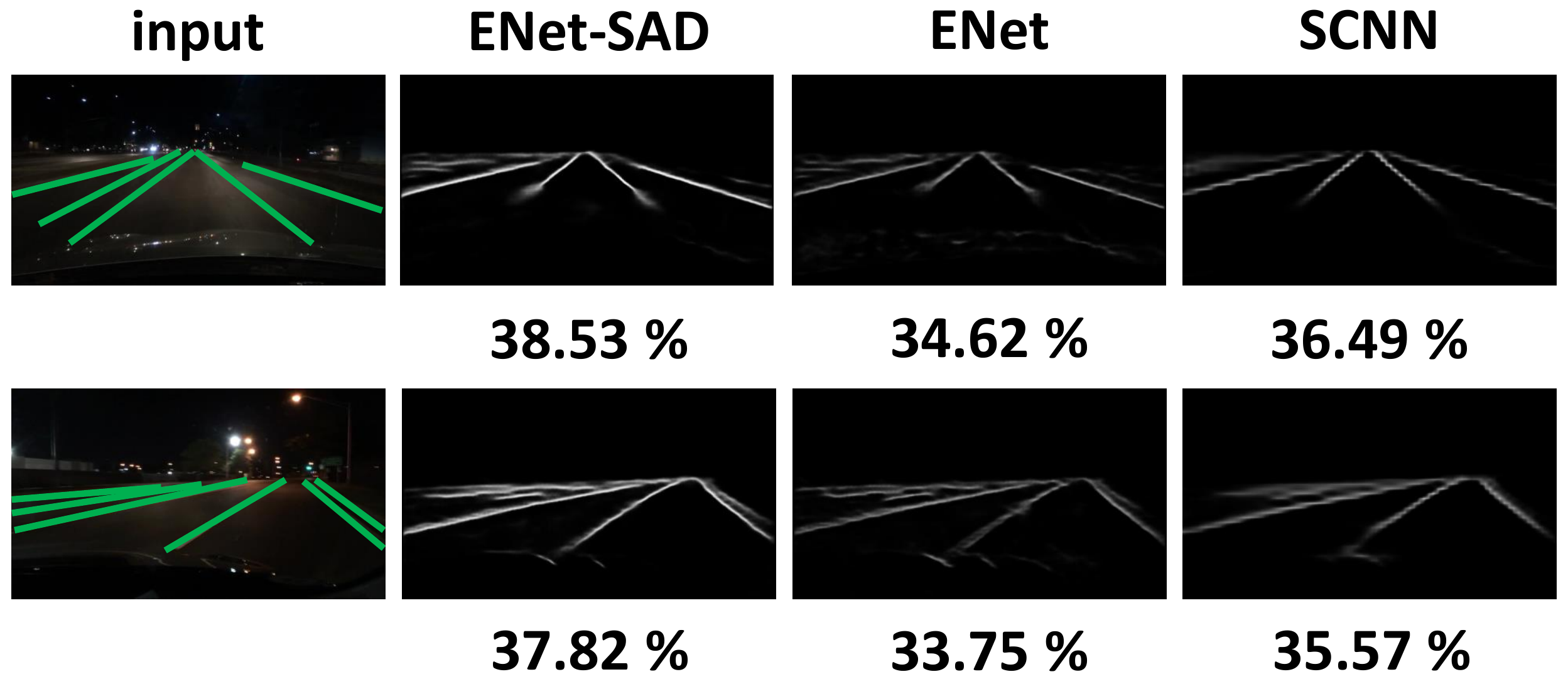}
  \vskip -0.2cm
  \caption{Performance of different algorithms on BDD100K testing set. We visualize the probability maps to better showcase the effect of adding self attention distillation. The brightness of the pixel indicates the probability of this pixel belonging to lanes. The number below each image denotes the pixel accuracy of lanes. Ground-truth lanes are drawn on the input image.}
  \centering
  \vskip -0.5cm
  \label{fig:bdd100k_result}
\end{figure}

We show some qualitative results of our algorithm and some baselines in these three benchmarks. As can be seen in Fig.~\ref{fig:tusimple_result}, ENet-\algorithmname~can detect lanes more precisely than ENet~\cite{paszke2016enet} in TuSimple and CUlane. 
As can be seen in Fig. ~\ref{fig:bdd100k_result}, the output probability maps of ENet-\algorithmname~are more compact and contain less noise compared with those of vanilla ENet and SCNN in poor lighting conditions. However, since many images in BDD100K contain more than 8 lanes and are collected in challenging scenarios like severe occlusion and poor lighting conditions, the performance of all algorithms is unsatisfactory and needs further improvement. In general, \algorithmname~can improve the visual attention as well as the detection performance in challenging conditions like crowded roads and poor light conditions. 

We also perform experiments that apply SAD and remove the effect of the P1 branch by blocking the gradient of the P1 branch from the main branch. Results show that ENet-SAD (without supervision from P1 branch) can still achieve 96.61$\%$ on TuSimple, 70.8 on CULane and 36.54$\%$ on BDD100K, which means the performance gains come mainly from \algorithmname~itself.

\subsection{Ablation Study}

We investigate the effects of different factors, e.g., the mimicking path, on the final performance. Besides, we also perform extensive experiments to investigate the timepoint to introduce \algorithmname~in the training process.


\noindent \textbf{Distillation paths of \algorithmname.} We summarize the performance of performing \algorithmname~between different blocks of ENet in Table \ref{attention_ablation_table}. We have a few observations. (1) \algorithmname~works well in the middle and high-level layers. (2) Adding \algorithmname~in low level layers will degrade the performance. 
The reason why \algorithmname~does not work in low-level layers is that these layers are originally designated to detect low-level details of the scene. Making them to mimic the attention maps of later layers will inevitably harm their ability of detecting local features since later layers encode more global information. 
Besides, we also find that mimicking the attention maps of the neighbouring layer successively brings more performance gains compared with mimicking those of non-adjacent layers ($P_{23}$ + $P_{34}$ outperforms $P_{24}$ + $P_{34}$). 
We conjecture that attention maps of neighbouring layers are closer from the semantic perspective compared with those of non-neighbouring layers (see Fig.~\ref{fig:attention_map}).  

\noindent \textbf{Backward distillation.} We also tested another distillation scheme that makes higher layers to mimic lower layers. It decreases the performance of ENet from 93.02\% to 91.26\% in TuSimple dataset. This is not surprising as low-level attention maps contain more details and are more noisy. Having higher-level layers to mimic lower layers will inevitably interfere the global information captured in higher layers, hampering the crucial clues for the lane detection task.
 
\begin{table}[!t]
\caption{Performance of different variants of ENet-\algorithmname~on TuSimple testing set. $P_{ij}$ denotes that the output of the $i$-$th$ block of ENet mimics the output of the $j$-th block.}
\label{attention_ablation_table}
\centering
\small{
\begin{tabular}{c|c||c|c||c|c}
\hline
Path & Accuracy & Path & Accuracy & Path & Accuracy \\
\hline \hline
$P_{12}$ & 91.22\% & $P_{23}$ & 94.72\% & $P_{23}$, $P_{24}$ & 95.38\% \\
$P_{13}$ & 91.36\% & $P_{24}$ & 94.63\% & \textbf{$P_{23}$, $P_{34}$} & \textbf{96.64\%} \\
$P_{14}$ & 91.47\% & $P_{34}$ & 95.29\% & $P_{24}$, $P_{34}$ & 96.52\% \\
\hline
\end{tabular}
}
\vspace{-2ex}
\end{table}

\noindent \textbf{\algorithmname~ v.s. Deep Supervision.} We also compare \algorithmname~with deep supervision~\cite{Xie_2015_ICCV}. Here, deep supervision denotes the algorithm that uses the labels directly as supervision for each layer in the network. More specifically, we use 1x1 convolution and bilinear upsampling to obtain the prediction of intermediate layers and use the cross entropy loss to train the intermediate outputs of the model. We empirically find that adding deep supervision in blocks 2 to 4 obtains the most significant performance gains. 
As can be seen in Table~\ref{deep_ablation_table}, \algorithmname~brings more performance gains than deep supervision in all three benchmarks. 
We attribute this to the following reasons. Firstly, compared with labels that are considered sparse and rigid, \algorithmname~provides softer attention targets that capture more contextual information that indicate the 
scene structure. Distilling information from attention maps of later layers helps previous layers to grasp the contextual signals. Secondly, a SAD path offers a feedback connection from deeper layers to shallower layers. The connection helps facilitate reciprocal learning between successive layers through attention distillation.

\begin{table}[!t]
\caption{Performance of \algorithmname~and deep supervision applied to different base models on TuSimple, CULane and BDD100K testing sets.}
\label{deep_ablation_table}
\centering
\small{
\begin{tabular}{c|c|c|c|c}
\hline
\multirow{2}*{Algorithm} & TuSimple & CULane & \multicolumn{2}{|c}{BDD100K} \\
\cline{2-5}
~ & Accuracy & Total & Accuracy & IoU \\
\hline \hline
ENet & 93.02\% & 68.4 & 34.12\% & 14.64 \\
ENet-Deep & 94.69\% & 69.6 & 35.61\% & 15.38 \\
ENet-\algorithmname & \textbf{96.64\%} & \textbf{70.8} & \textbf{36.56\%} & \textbf{16.02} \\
\hline
R-18 & 92.69\% & 67.9 & 30.66\% & 11.07 \\
R-18-Deep & 94.14\% & 68.8 & 30.95\% & 12.23 \\
R-18-\algorithmname & \textbf{96.02\%} & \textbf{70.5} & \textbf{31.10\%} & \textbf{13.29} \\
\hline
R-34 & 92.84\% & 68.6 & 30.92\% & 12.24 \\
R-34-Deep & 94.52\% & 69.2 & 31.72\% & 13.59 \\
R-34-\algorithmname & \textbf{96.24\%} & \textbf{70.7} & \textbf{32.68\%} & \textbf{14.56} \\
\hline
\end{tabular}
}
\vspace{-3ex}
\end{table}

\begin{figure}[t]
  \centering
  \includegraphics[width=0.8\linewidth]{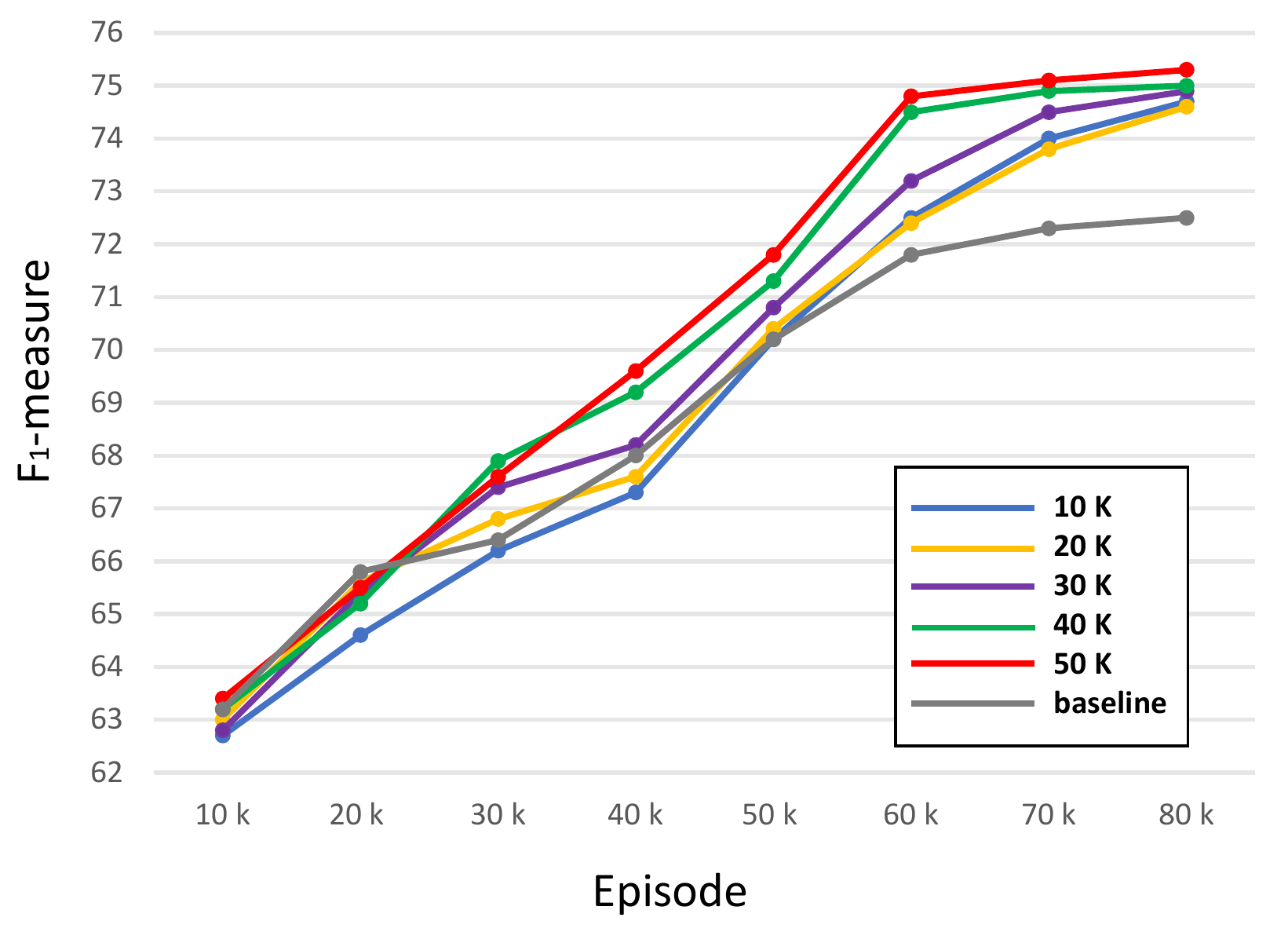}
  \vskip -0.2cm
  \caption{Performance of adding self attention distillation on the ENet model at different training episodes on the CULane validation set. The number in the legend denotes the episode when self attention distillation is added. "Baseline" denotes the ENet model without self attention distillation.}
  \vskip -0.1cm
  \centering
  \label{fig:when_add_SAD}
\end{figure}

\noindent \textbf{When to add \algorithmname.} Recall that we assume a half-trained model before we add SAD into the training. Here, we investigate the different timepoints to add \algorithmname. As can be seen in Fig.~\ref{fig:when_add_SAD}, although different timepoints of introducing SAD  achieve almost the same performance in the end, the time to add \algorithmname~ has an effect on the convergence speed of the networks. We attribute the phenomenon to the quality of the target attention maps produced by later layers. In earlier training stage, deeper layers have not been trained well and therefore the distillation targets produced by these layers are of low quality. Introducing SAD at these earlier stages is not as fruitful. Conversely, adding SAD in later training stage would benefit the representation learning of the previous layers. 

\section{Discussion}\label{conclusion}

We have proposed a simple yet effective attention distillation approach, \ie, \algorithmname, to improve the representation learning of CNN-based lane detection models. \algorithmname~is validated in various models (\ie, ENet, ResNet-18, ResNet-34, and ResNet-101) and achieves consistent performance gains in three popular benchmarks (\ie, TuSimple, CULane and BDD100K), demonstrating the effectiveness of \algorithmname. The results show that \algorithmname~can generally improve the visual attention of different layers in various networks. 
It would be interesting to extend this idea to other tasks that demands fine-grained attention to details, such as image saliency detection and image matting. 

%

\noindent
\textbf{Acknowledgement:} This work is supported by SenseTime Group Limited, the General Research Fund sponsored by the Research Grants Council of the Hong Kong SAR (CUHK 14241716), Singapore MOE AcRF Tier 1 (M4012082.020), NTU SUG, and NTU NAP.

{\small
\bibliographystyle{ieee}
\bibliography{egbib}
}

\clearpage

\appendix

\begin{figure*}[!ht]
  \centering
  \includegraphics[width=1.0\linewidth]{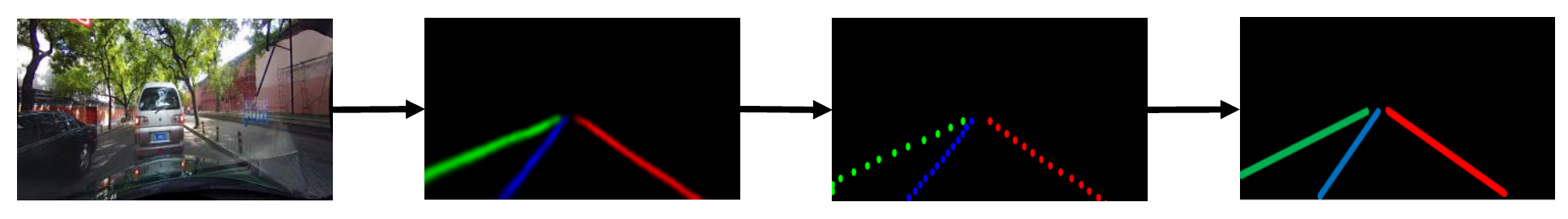}
  \vskip -0.2cm
  \caption{The process of obtaining lanes from probability maps on the CULane dataset. From left to right: original image, probability map, extracted lane points and final lane prediction.}
  \centering
  \label{fig:show_point}
\end{figure*}

\section{Details of Architecture}

Table~\ref{exist_table} summarizes the architecture of the lane existence prediction branch for ENet-\algorithmname, ResNet-18-\algorithmname~and ResNet-34-\algorithmname. As to ResNet-18-\algorithmname~and ResNet-34-\algorithmname, we also use dilated convolution~\cite{yu2015multi} to replace the original convolution layers in the last two blocks for ResNet-18~\cite{he2016deep} and ResNet-34~\cite{he2016deep}. 

\begin{table}[h]
\caption{The architecture of the the lane existence prediction branch. Assuming the input is 3 $\times$ 288 $\times$ 800. Note that the output size is $c$ $\times$ $h$ $\times$ $w$ before "Flatten", where $c$, $h$ and $w$ denote channel, height and width, respectively. The number in the bracket besides the layer name is the parameter for that layer. For instance, the four numbers besides dilated convolution denote kernel size, stride, padding and dilated rate, respectively.}
\vskip 0.1cm
\label{exist_table}
\centering
\small{
\begin{tabular}{l|c}
\hline
Layer Name & Output Size \\
\hline
\hline
Dilated Convolution (3, 1, 4, 4) & 32 $\times$ 36 $\times$ 100 \\
Batch Normalization & 32 $\times$ 36 $\times$ 100 \\
Relu & 32 $\times$ 36 $\times$ 100 \\
Spatial Dropout (0.1) & 32 $\times$ 36 $\times$ 100 \\
Convolution (1, 1) & 5 $\times$ 36 $\times$ 100 \\
Spatial SoftMax & 5 $\times$ 36 $\times$ 100 \\
Average Pooling & 5 $\times$ 18 $\times$ 50 \\
\hline
Flatten & 4500 \\
Fully Connected & 128 \\
Relu & 128 \\
Fully Connected & 4 \\
Sigmoid & 4 \\
\hline
\end{tabular}
}
\end{table}

\section{Lane Post-processing in CULane}

For CULane, in the inference stage, we feed the image into the ENet model. Then the multi-channel probability maps and the lane existence vector are obtained. Following~\cite{pan2017spatial}, the final output is obtained as follows: First, we use a 9 $\times$ 9 kernel to smooth the probability maps. Then, for each lane whose existence probability is larger than 0.5, we search the corresponding probability map every 20 rows for the position with the highest probability value. Finally, we use cubic splines to connect these positions to get the final output. The process improves the final lane prediction results as it removes noises in the probability maps.
The process is depicted in Figure~\ref{fig:show_point}. Here, we differentiate different lane instances with different colors. 

\section{More Qualitative Results in Lane Detection}

\begin{figure}[t]
  \centering
  \includegraphics[width=\linewidth]{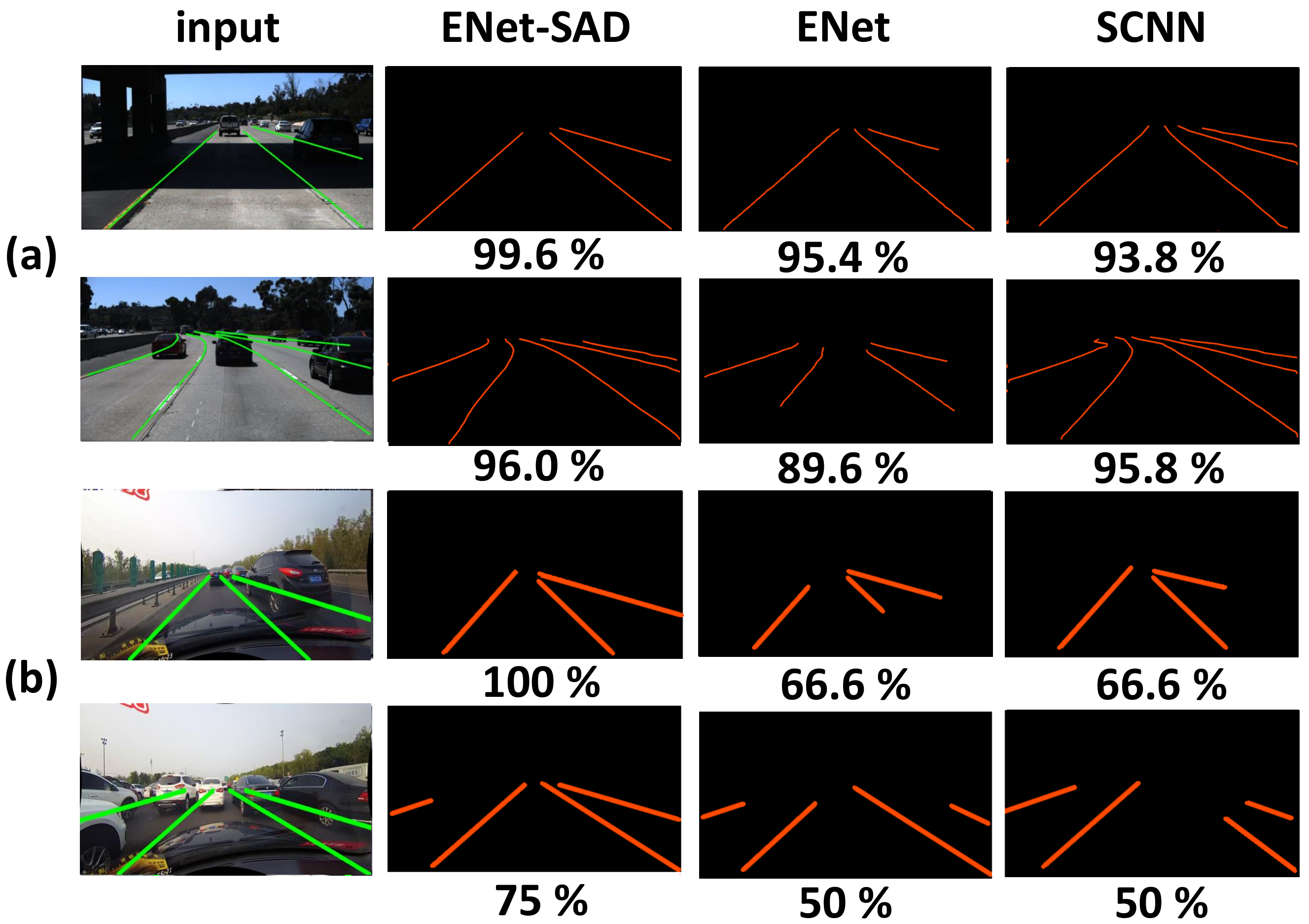}
  \vskip -0.1cm
  \caption{Performance of different algorithms on (a) TuSimple and (b) CULane testing sets. The number below each image denotes the accuracy. Ground-truth lanes are drawn on the input image.}
  \centering
  \label{fig:tusimple_result_supp}
\end{figure}

\begin{figure}[t]
  \centering
  \includegraphics[width=\linewidth]{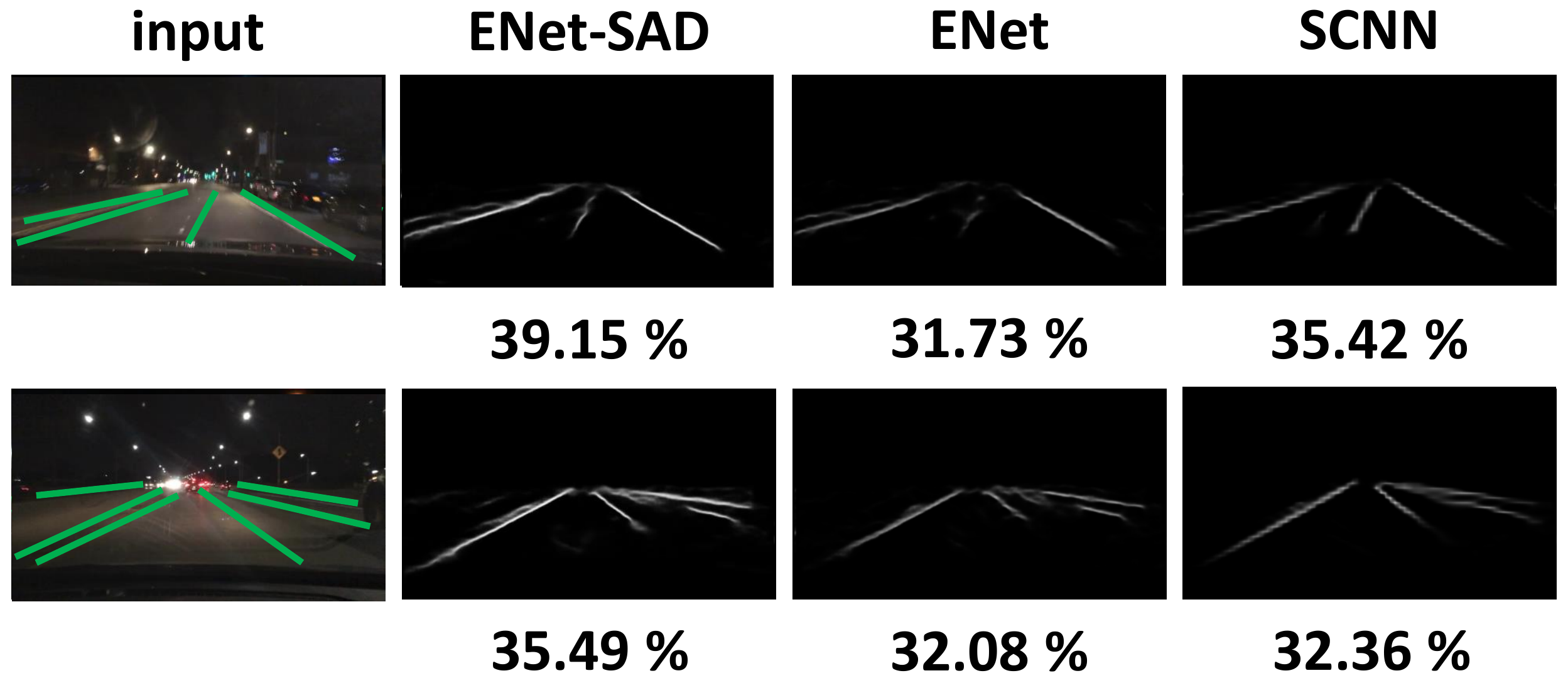}
  \vskip -0.2cm
  \caption{Performance of different algorithms on BDD100K testing set. We visualize the probability maps to better showcase the effect of adding self attention distillation. The brightness of the pixel indicates the probability of this pixel belonging to lanes. The number below each image denotes the pixel accuracy of lanes. Ground-truth lanes are drawn on the input image.}
  \centering
  \label{fig:bdd100k_result_supp}
\end{figure}

Figures~\ref{fig:tusimple_result_supp} and ~\ref{fig:bdd100k_result_supp} depict the qualitative results of different algorithms on TuSimple~\cite{tusimple}, CULane~\cite{pan2017spatial} and BDD100K~\cite{yu2018bdd100k}. As can be seen in Fig.~\ref{fig:tusimple_result_supp}, ENet-\algorithmname~can detect lanes more precisely than ENet~\cite{paszke2016enet} in TuSimple and CUlane. Besides, the detection of ENet-\algorithmname~is less affected by the irrelevant objects on the road compared with SCNN~\cite{pan2017spatial}. As can be seen in Fig.~\ref{fig:bdd100k_result_supp}, the output probability maps of ENet-\algorithmname~are more compact and contain less noise compared with those of SCNN in poor light conditions.

{\small
\bibliographystyle{ieee}
\bibliography{egbib}

\begin{thebibliography}{10}\itemsep=-1pt

\bibitem{bertozzi1998gold}
M.~Bertozzi and A.~Broggi.
\newblock Gold: A parallel real-time stereo vision system for generic obstacle
  and lane detection.
\newblock {\em IEEE Transactions on Image Processing}, 7(1):62--81, 1998.

\bibitem{borkar2012novel}
A.~Borkar, M.~Hayes, and M.~T. Smith.
\newblock A novel lane detection system with efficient ground truth generation.
\newblock {\em IEEE Transactions on Intelligent Transportation Systems},
  13(1):365--374, 2012.

\bibitem{bottou2010large}
L.~Bottou.
\newblock Large-scale machine learning with stochastic gradient descent.
\newblock In {\em International Conference on Computational Statistics}, pages
  177--186. Springer, 2010.

\bibitem{chen2016attention}
L.-C. Chen, Y.~Yang, J.~Wang, W.~Xu, and A.~L. Yuille.
\newblock Attention to scale: scale-aware semantic image segmentation.
\newblock In {\em IEEE Conference on Computer Vision and Pattern Recognition},
  pages 3640--3649, 2016.

\bibitem{chen2017rbnet}
Z.~Chen and Z.~Chen.
\newblock Rbnet: A deep neural network for unified road and road boundary
  detection.
\newblock In {\em International Conference on Neural Information Processing},
  pages 677--687. Springer, 2017.

\bibitem{deusch2012random}
H.~Deusch, J.~Wiest, S.~Reuter, M.~Szczot, M.~Konrad, and K.~Dietmayer.
\newblock A random finite set approach to multiple lane detection.
\newblock In {\em IEEE Conference on Intelligent Transportation Systems}, pages
  270--275. IEEE, 2012.

\bibitem{furlanello2018born}
T.~Furlanello, Z.~Lipton, M.~Tschannen, L.~Itti, and A.~Anandkumar.
\newblock Born-again neural networks.
\newblock In {\em International Conference on Machine Learning}, pages
  1602--1611, 2018.

\bibitem{ghafoorian2018gan}
M.~Ghafoorian, C.~Nugteren, N.~Baka, O.~Booij, and M.~Hofmann.
\newblock {EL-GAN}: embedding loss driven generative adversarial networks for
  lane detection.
\newblock In {\em European Conference on Computer Vision}, pages 256--272.
  Springer, 2018.

\bibitem{he2016accurate}
B.~He, R.~Ai, Y.~Yan, and X.~Lang.
\newblock Accurate and robust lane detection based on dual-view convolutional
  neutral network.
\newblock In {\em IEEE Intelligent Vehicles Symposium}, pages 1041--1046. IEEE,
  2016.

\bibitem{he2016deep}
K.~He, X.~Zhang, S.~Ren, and J.~Sun.
\newblock Deep residual learning for image recognition.
\newblock In {\em IEEE Conference on Computer Vision and Pattern Recognition},
  pages 770--778, 2016.

\bibitem{hinton2015distilling}
G.~Hinton, O.~Vinyals, and J.~Dean.
\newblock Distilling the knowledge in a neural network.
\newblock {\em STAT}, 1050:9, 2015.

\bibitem{hou2018learning}
Y.~Hou, Z.~Ma, C.~Liu, and C.~C. Loy.
\newblock Learning to steer by mimicking features from heterogeneous auxiliary
  networks.
\newblock In {\em Association for the Advancement of Artificial Intelligence},
  2019.

\bibitem{jaderberg2015spatial}
M.~Jaderberg, K.~Simonyan, A.~Zisserman, et~al.
\newblock Spatial transformer networks.
\newblock In {\em Advances in Neural Information Processing Systems}, pages
  2017--2025, 2015.

\bibitem{lee2017vpgnet}
S.~Lee, J.~Kim, J.~S. Yoon, S.~Shin, O.~Bailo, N.~Kim, T.-H. Lee, H.~S. Hong,
  S.-H. Han, and I.~S. Kweon.
\newblock Vpgnet: Vanishing point guided network for lane and road marking
  detection and recognition.
\newblock In {\em IEEE International Conference on Computer Vision}, pages
  1965--1973. IEEE, 2017.

\bibitem{neven2018towards}
D.~Neven, B.~De~Brabandere, S.~Georgoulis, M.~Proesmans, and L.~Van~Gool.
\newblock Towards end-to-end lane detection: an instance segmentation approach.
\newblock In {\em IEEE Intelligent Vehicles Symposium}, pages 286--291. IEEE,
  2018.

\bibitem{pan2017spatial}
X.~Pan, J.~Shi, P.~Luo, X.~Wang, and X.~Tang.
\newblock Spatial as deep: Spatial {CNN} for traffic scene understanding.
\newblock In {\em Association for the Advancement of Artificial Intelligence},
  2018.

\bibitem{paszke2016enet}
A.~Paszke, A.~Chaurasia, S.~Kim, and E.~Culurciello.
\newblock {ENet}: A deep neural network architecture for real-time semantic
  segmentation.
\newblock {\em arXiv preprint arXiv:1606.02147}, 2016.

\bibitem{tusimple}
TuSimple.
\newblock http://benchmark.tusimple.ai/\#/t/1.
\newblock Accessed: 2018-09-08.

\bibitem{Wang_2017_CVPR}
F.~Wang, M.~Jiang, C.~Qian, S.~Yang, C.~Li, H.~Zhang, X.~Wang, and X.~Tang.
\newblock Residual attention network for image classification.
\newblock In {\em IEEE Conference on Computer Vision and Pattern}, 2017.

\bibitem{Xie_2015_ICCV}
S.~Xie and Z.~Tu.
\newblock Holistically-nested edge detection.
\newblock In {\em IEEE International Conference on Computer Vision}, 2015.

\bibitem{yim2017gift}
J.~Yim, D.~Joo, J.~Bae, and J.~Kim.
\newblock A gift from knowledge distillation: Fast optimization, network
  minimization and transfer learning.
\newblock In {\em IEEE Conference on Computer Vision and Pattern Recognition},
  pages 4133--4141, 2017.

\bibitem{yu2015multi}
F.~Yu and V.~Koltun.
\newblock Multi-scale context aggregation by dilated convolutions.
\newblock In {\em International Conference on Learning Representations}, 2016.

\bibitem{yu2018bdd100k}
F.~Yu, W.~Xian, Y.~Chen, F.~Liu, M.~Liao, V.~Madhavan, and T.~Darrell.
\newblock {BDD100K}: A diverse driving video database with scalable annotation
  tooling.
\newblock {\em arXiv preprint arXiv:1805.04687}, 2018.

\bibitem{zagoruyko2016paying}
S.~Zagoruyko and N.~Komodakis.
\newblock Paying more attention to attention: improving the performance of
  convolutional neural networks via attention transfer.
\newblock In {\em International Conference on Learning Representations}, 2017.

\bibitem{zhang2018geometric}
J.~Zhang, Y.~Xu, B.~Ni, and Z.~Duan.
\newblock Geometric constrained joint lane segmentation and lane boundary
  detection.
\newblock In {\em European Conference on Computer Vision}, pages 486--502,
  2018.

\end{thebibliography}
}

\end{document}